%% file: main.tex
\documentclass{article}

\usepackage{PRIMEarxiv}
\input{macro}

\input{math_commands}

\usepackage[utf8]{inputenc} %
\usepackage[T1]{fontenc}    %
\usepackage{hyperref}       %
\usepackage{url}            %
\usepackage{booktabs}       %
\usepackage{amsfonts}       %
\usepackage{nicefrac}       %
\usepackage{microtype}      %
\usepackage{fancyhdr}       %

\newcommand{\ourmethod}{\emph{D-Cubed }}

\title{D-Cubed: Latent Diffusion Trajectory Optimisation for Dexterous Deformable Manipulation}

\author{
  Jun Yamada, Shaohong Zhong, Jack Collins, Ingmar Posner \\
  Applied A2I Lab \\
  University of Oxford \\
  \texttt{\{jyamada, shaohong, jcollins, ingmar\}@robots.ox.ac.uk} \\
}

\begin{document}
\maketitle
\input{sections/abstract}


\input{sections/introduction}
\input{sections/related_works}
\input{sections/preliminary}
\input{sections/approach}

\input{sections/experiments}

\input{sections/conclusion}

\section*{Acknowledgments}
This work was supported by a UKRI/EPSRC Programme Grant [EP/V000748/1]. We would like to thank Anson Lei and Frederik Nolte for their valuable feedback. We would also like to acknowledge the use of the University of Oxford Advanced Research Computing (ARC) facility in carrying out this work.

\bibliographystyle{unsrt}  
\bibliography{references}  

\clearpage
\input{sections/appendix}

\end{document}

%% file: macro.tex
\usepackage{algorithm}
\usepackage[noend]{algpseudocode}
\usepackage{float}
\usepackage{bbm}
\usepackage{graphicx}
\usepackage{subcaption}
\usepackage{hyperref}       %
\usepackage{url}            %
\usepackage{booktabs}       %
\usepackage{amsfonts}       %
\usepackage{amsmath,amssymb}
\usepackage{nicefrac}       %
\usepackage{amsmath,amsfonts,bm}
\usepackage{enumitem}
\usepackage{wrapfig}
\usepackage{mathtools}
\usepackage[table]{xcolor}
\usepackage{multirow}
\usepackage{listings}

\newcommand{\Skip}[1]{}

\newcommand{\ie}{i.e.,\ }

\newcommand{\RNum}[1]{\uppercase\expandafter{\romannumeral #1\relax}}

\definecolor{codegreen}{rgb}{0,0.6,0}
\definecolor{codegray}{rgb}{0.5,0.5,0.5}
\definecolor{codepurple}{rgb}{0.58,0,0.82}
\definecolor{backcolour}{rgb}{0.95,0.95,0.92}

\lstdefinestyle{mystyle}{
    backgroundcolor=\color{backcolour},   
    commentstyle=\color{codegreen},
    keywordstyle=\color{magenta},
    numberstyle=\tiny\color{codegray},
    stringstyle=\color{codepurple},
    basicstyle=\ttfamily\footnotesize,
    breakatwhitespace=false,         
    breaklines=true,                 
    captionpos=b,                    
    keepspaces=true,                 
    numbers=left,                    
    numbersep=5pt,                  
    showspaces=false,                
    showstringspaces=false,
    showtabs=false,                  
    tabsize=2
}

%% file: math_commands.tex
\def\eqref#1{equation~\ref{#1}}

\def\1{\bm{1}}

\DeclareMathAlphabet{\mathsfit}{\encodingdefault}{\sfdefault}{m}{sl}
\SetMathAlphabet{\mathsfit}{bold}{\encodingdefault}{\sfdefault}{bx}{n}

%% file: sections/abstract.tex
\begin{abstract}

Mastering dexterous robotic manipulation of deformable objects is vital for overcoming the limitations of parallel grippers in real-world applications.
Current trajectory optimisation approaches often struggle to solve such tasks due to the large search space and the limited task information available from a cost function. 
In this work, we propose \ourmethod, a novel trajectory optimisation method using a latent diffusion model (LDM) trained from a task-agnostic play dataset to solve dexterous deformable object manipulation tasks.
\ourmethod learns a skill-latent space that encodes short-horizon actions in the play dataset using a VAE and trains a LDM to compose the skill latents into a skill trajectory, representing a long-horizon action trajectory in the dataset.
To optimise a trajectory for a target task, we introduce a novel gradient-free guided sampling method that employs the Cross-Entropy method within the reverse diffusion process.
In particular, \ourmethod samples a small number of noisy skill trajectories using the LDM for exploration and evaluates the trajectories in simulation. 
Then \ourmethod selects the trajectory with the lowest cost for the subsequent reverse process.
This effectively explores promising solution areas and optimises the sampled trajectories towards a target task throughout the reverse diffusion process.
Through empirical evaluation on a public benchmark of dexterous deformable object manipulation tasks, we demonstrate that \ourmethod outperforms traditional trajectory optimisation and competitive baseline approaches by a significant margin. We further demonstrate that trajectories found by \ourmethod readily transfer to a real-world LEAP hand on a folding task.

\end{abstract}

%% file: sections/introduction.tex
\section{Introduction}
The realm of dexterous robot hand manipulation has made remarkable progress in recent years, in part due to advances in learning-based methods~\cite{andrychowicz2020learning, pmlr-v164-chen22a, nagabandi2020deep}.
However, past research has focused predominantly on tasks that involve rigid objects~\cite{mordatch2012contact, bai2014dexterous, sundaralingam2018relaxedrigidity, charlesworth2021solving}.
On the other hand, real-world manipulation tasks often present scenarios in which robots need to manipulate deformable objects, such as folding a piece of clothing~\cite{Tsurumine2018cloth}, manipulating soft tissues~\cite{pore2021learning} or dough~\cite{shi2022robocraft, shi2023robocook}.

One common approach to generating actions for a dexterous robot hand is trajectory optimisation, which optimises an action sequence by minimising a task-informed cost function. 
However, the application of trajectory optimisation is predominantly limited to rigid object manipulation~\cite{charlesworth2021solving} or relatively simple deformable object manipulation tasks with short horizons~\cite{huang2021plasticinelab}.
The primary challenges of optimising a trajectory for complex tasks such as those seen in dexterous deformable object manipulation stem from 1) the large search space due to the complexity of the task including the infinite dimensionality of deformable objects and high degree-of-freedom (DoF) of the robot hand; 2) the large number of contacts associated with handling the objects; 3) the limited task information that the cost function typically provides~\cite{antonova2022rethinking}. 
Commonly, the cost function to be optimised is defined as the distance between a target shape and the final shape of a deformable object~\cite{li2023dexdeform}.
In this scenario, task-relevant signal is typically unavailable when no contact is made between the robot and the manipulated object, inhibiting the optimisation of a feasible trajectory.
While prior work~\cite{li2023dexdeform} utilises expert demonstrations with trajectory optimisation to alleviate this issue, collecting such demonstrations for each task is considered expensive.

\begin{wrapfigure}{r}{0.5\textwidth}
    \vspace{-0.4cm}
    \centering
    \includegraphics[width=0.45\textwidth]{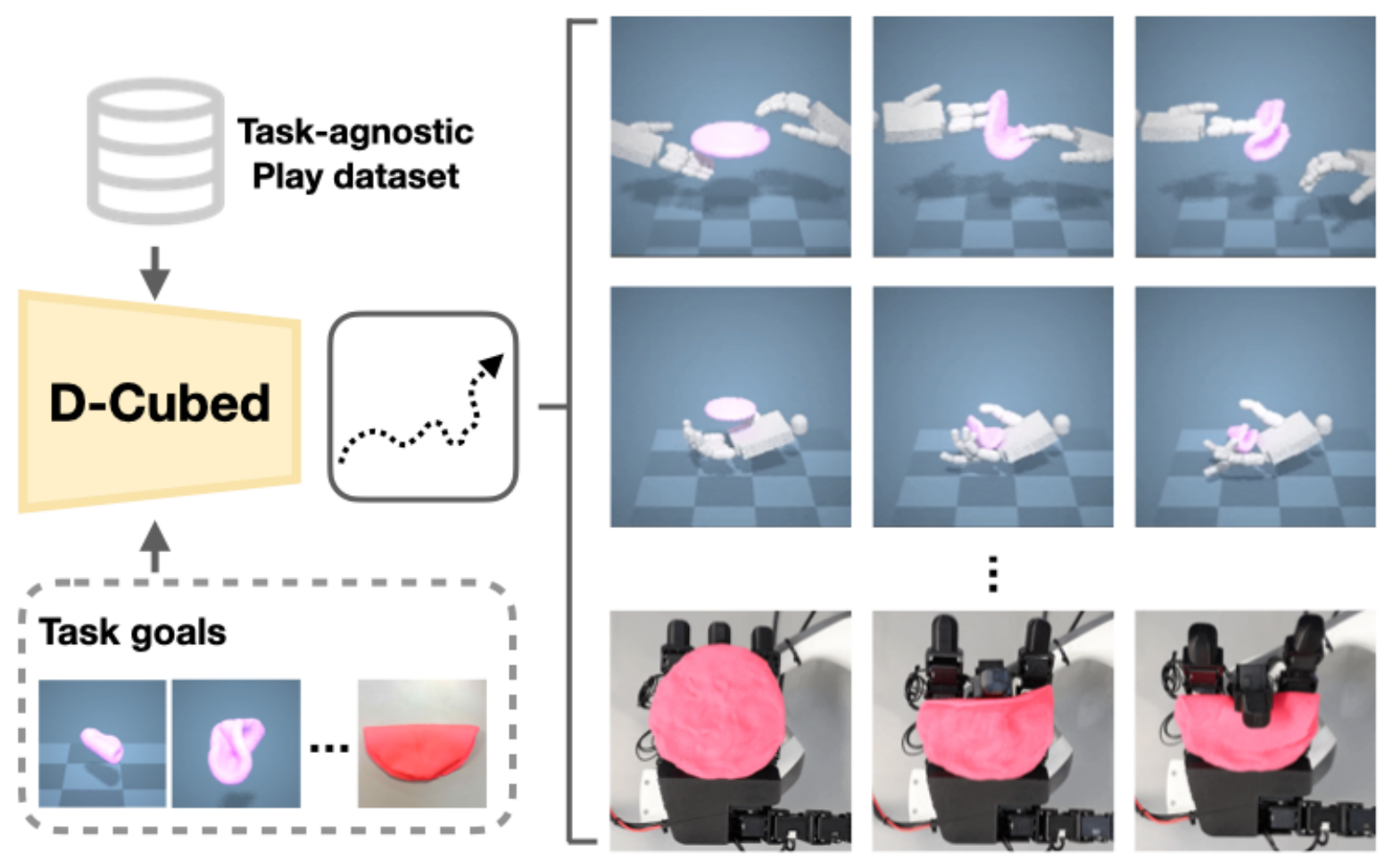}
    \caption{\ourmethod leverages a latent diffusion model trained from a task-agnostic play dataset to generate open-loop action trajectories for long-horizon dexterous deformable object manipulation tasks.}
    \label{fig:teaser}
    \vspace{-0.2cm}
\end{wrapfigure}

In this work, we propose \emph{D-Cubed}, Latent \textbf{D}iffusion for Trajectory Optimisation in \textbf{D}exterous \textbf{D}eformable Manipulation (see Fig~\ref{fig:teaser}). 
\ourmethod is a novel trajectory optimisation approach that leverages a latent diffusion model (LDM)~\cite{ho2020denoising} trained on a task-agnostic play dataset of a robot hand that contains various representative hand motions, such as closing and opening the hand, and moving individual fingers.
In particular, \ourmethod learns a skill-latent space that encodes short-horizon action sequences from the play dataset using a variational autoencoder (VAE). 
An LDM is then trained to compose these skill-latent representations into a skill trajectory that corresponds to the motions of the robot hand seen in the dataset.
The LDM, capable of generating diverse trajectories of meaningful robot hand motions, is applicable across diverse manipulation tasks.
To find a performant action trajectory for a target task, we propose a novel gradient-free guided sampling method that employs a variation of the Cross-Entropy Method (CEM)~\cite{RUBINSTEIN199789, marin2012ce} within the reverse diffusion process.
For each denoising step, the LDM generates a small number of noisy skill-latent trajectories to explore the solution space.
Since the skill latent space is trained to represent smooth low-level action sequences, each skill in these noisy skill trajectories produces meaningful and consistent action trajectories that facilitate efficient exploration.
These skill-latent trajectories are evaluated in a simulator using a predefined cost function, and the trajectory with the lowest task cost among the sampled trajectories is selected for further denoising in the reverse diffusion process.
With each denoising step guided by the CEM, the noise in the chosen performant trajectory is gradually removed, refining it towards solutions with lower costs.

In summary, our contributions are three-fold: (1) we propose \ourmethod, a trajectory optimisation method using an LDM to solve challenging long-horizon dexterous manipulation tasks; 
(2) we introduce a novel gradient-free guided sampling method that employs the CEM within the reverse diffusion process to optimise a trajectory for a target task;
and (3) we empirically demonstrate that \ourmethod significantly outperforms competitive baselines, including traditional trajectory optimisation methods such as gradient-based and sampling-based approaches.

%% file: sections/related_works.tex
\section{Related Works}
Several prior works note the importance of deformable object manipulation and introduce benchmark tasks for evaluating competing methodologies~\cite{huang2021plasticinelab, lin2021softgym, chen2022benchmarking, blancomulero2024benchmarking, li2023dexdeform}.
While most benchmarks focus on deformable object manipulation tasks with point-mass agents or parallel grippers that are incapable of dexterous manipulation, \cite{li2023dexdeform} propose a suite of deformable object manipulation tasks with dexterous robot hands~\cite{shadowhand} built upon a differentiable physics engine.

Trajectory optimisation, including gradient-based and sampling-based, is a common approach to solving dexterous robot hand manipulation tasks by assuming access to an accurate dynamics model or assuming simplified object geometries (e.g.~\cite{mordatch2012contact, bai2014dexterous, sundaralingam2018relaxedrigidity, charlesworth2021solving}). 
Gradient-based approaches directly optimise a task-informed cost function through a learnt dynamics model~\cite{nagabandi2020deep, kumar2016optimal, yamada2023leveraging} or differentiable simulator~\cite{huang2021plasticinelab, li2023dexdeform, qiao2021differentiable, hu2019difftaichi} to find a performant action sequence.
However, the application of gradient-based trajectory optimisation is predominantly limited to relatively simple tasks with a short horizon~\cite{huang2021plasticinelab}. 
This is because gradients available from the environment do not often include global task information leading to locally optimal solutions, further exacerbated by the large number and nonlinearity of contacts~\cite{li2023dexdeform, lin2022diffskill}.

Sampling-based methods such as the CEM~\cite{RUBINSTEIN199789,marin2012ce} and Model Predictive Path Integrals (MPPI)~\cite{williams2017mppi} offer a simple and effective approach to optimising an action trajectory by sampling actions for exploration.
In particular, MPPI~\cite{williams2017mppi} has been applied to dexterous hand manipulation tasks with rigid objects~\cite{charlesworth2021solving}.
However, such sampling-based methods tend to be computationally expensive when the solution space is large because they require sampling a large number of trajectories to effectively search the solution space.

To alleviate such issues, prior works train 
policies from expert demonstrations~\cite{Ze2024DP3} or combine expert demonstrations with trajectory optimisation~\cite{li2023dexdeform} for dexterous deformable object manipulation.
However, collecting expert demonstrations for each new task is considered expensive.
Notably, rather than collecting expert demonstrations, in this work, a task-agnostic play dataset that contains representative movements of the hand is collected to form a structured skill-latent space that is used across diverse tasks.

Reinforcement learning (RL) is an alternative approach to trajectory optimisation that does not require expert demonstrations.
Both model-based RL~\cite{nagabandi2019deep} and model-free RL~\cite{rajeswaran2018learning, OpenAI2019SolvingRC} have shown promising results for rigid object manipulation tasks with a dexterous robot hand.
However, RL typically requires a large number of samples for training, which is further exacerbated for deformable object manipulation tasks that feature infinite degrees of freedom.

Diffusion models, a class of generative models, formulate data generation as an iterative denoising process~\cite{pmlr-v37-sohl-dickstein15, ho2020denoising, song2020score}.
Classifier guidance~\cite{dhariwal2021diffusion} is a common technique for using gradients to guide the sampling process of unconditional diffusion models to generate a desired sample, including a trajectory for a target task~\cite{chi2023diffusion, janner2022diffuser, liang2023adaptdiffuser, rigter2023world}.
However, classifier guidance struggles to guide sampling when gradients are inaccurate~\cite{chao2022denoising}, such as when gradients are obtained from differentiable physics simulators~\cite{antonova2022rethinking}.
In contrast, this work proposes gradient-free guidance that employs a variation of CEM to the reverse diffusion process for trajectory optimisation.

Concurrent research~\cite{yang2024diffusiones} also proposes a gradient-free method based on evolutionary search for trajectory optimisation, evaluated on simple simulated driving tasks with well-engineered cost functions.
Their sampling method~\cite{yang2024diffusiones} initially completes the reverse diffusion process to generate clean trajectories from noisy samples. 
Then, for the mutation step of the evolutionary search, adds and then removes noise by taking short forward and then reverse denoising steps.
This method is unlikely to sufficiently cover large search spaces, because it does not drastically perturb the trajectories.
Crucially, their diffusion model is trained to generate low-level action trajectories instead of skills.
On the other hand, by directly using the CEM within the reverse diffusion process, \ourmethod focuses on exploration by sampling diverse noisy skill-latent trajectories at the beginning of the reverse process.
This allows \ourmethod to focus on exploration in the early diffusion steps and refine the trajectories in the later steps to overcome the exploration problems. %

%% file: sections/preliminary.tex
\section{Preliminaries}
\ourmethod employs a diffusion model to generate diverse action trajectories for exploration. 
\ourmethod also relies upon the CEM to optimise a trajectory to solve a target task within the reverse diffusion process.
In the following sections, we introduce the preliminaries of diffusion models (Section~\ref{sec:ddpm}) and CEM (Section~\ref{sec:cem}).

\subsection{Denoising Diffusion Probabilistic Models}
\label{sec:ddpm}

Denoising Diffusion Probabilistic Models (DDPMs)~\cite{ho2020denoising, sohl2015deep} are a class of generative models that are trained by denoising a sequence of noise-corrupted inputs.
For each training datum $\mathbf{x}_{0} \sim q_{data}(\mathbf{x})$, the forward diffusion process constructs a Markov chain $\mathbf{x}_0, \mathbf{x}_1, \dots, \mathbf{x}_N$ such that $q(\mathbf{x}_i | \mathbf{x}_{i-1}) = \mathcal{N}(\mathbf{x}_{i}; \sqrt{1-\beta_i} \mathbf{x}_{i-1}, \beta_{i}\mathbf{I})$
where $\beta_i$ denotes a positive noise scale and subscript indices $i$ refer to time steps of the diffusion process.

On the other hand, the reverse process, which aims to remove noise from the noisy sample $x_{i}$ is defined as $p_{\boldsymbol{\theta}}(\mathbf{x}_{0:N}) = p(\mathbf{x}_{N})\Pi^{N}_{i=1} p_{\boldsymbol{\theta}}(\mathbf{x}_{i-1} | \mathbf{x}_{i})$,
where $p(\mathbf{x}_{N})=\mathcal{N}(\mathbf{0}, \mathbf{I})$. The conditional distribution $p_{\boldsymbol{\theta}}(\mathbf{x}_{i-1} | \mathbf{x}_{i})$, commonly modelled as a Gaussian distribution with mean $\mu_{\boldsymbol{\theta}}(\mathbf{x}_{i}, i)$ and covariance $\Sigma_{\theta}(\mathbf{x}_{i}, i)$, is defined as $p_{\theta}(\mathbf{x}_{i-1} | \mathbf{x}_{i}) = \mathcal{N}(\mathbf{x}_{i-1}; \mu_{\theta}(\mathbf{x}_{i}, i), \Sigma_{\theta}(\mathbf{x}_{i}, i))$.
Given a sample $\mathbf{x}_{i}$, the mean $\mu_{\theta}(\mathbf{x}_{i}, i)$ and the variance $\Sigma_{\theta}(\mathbf{x}_{i}, i)$ are defined as:
\begin{equation}
\label{eq:mean}
    \mu_{\theta}(\mathbf{x}_{i}, i) = \frac{\sqrt{\bar{\alpha}_{i-1}} \beta_t}{1-\bar{\alpha}_i} \mathbf{x}_0+\frac{\sqrt{\alpha_t}\left(1-\bar{\alpha}_{i-1}\right)}{1-\bar{\alpha}_i} \mathbf{x}_i
\end{equation}
\begin{equation}
    \Sigma_{\boldsymbol{\theta}}(\mathbf{x}_i, i) = \sigma^2_{i}\mathbf{I} = \tilde{\beta}_{i} = \frac{1-\overline\alpha_i-1}{1-\overline\alpha_i}\beta_i,
    \label{eq:diffusion_variance}
\end{equation}
where $\alpha_{i} := 1-\beta_{i}$ and $\overline\alpha := \Pi^{i}_{j=1} \alpha_{j}$. 

Instead of predicting the noise, $\epsilon_{i}$~\cite{ho2020denoising}, added to the data, we train a diffusion mode $G_{\theta}(x_{i}, i)$ to directly predict the clean datapoint, $x_{0}$, to simplify the objective~\cite{ramesh2022hierarchical}:
\begin{equation}
    \mathcal{L}_{\text{diffusion}} = \mathbb{E}_{\mathbf{x}_{0}\sim q(\mathbf{x}_{0}), i\sim[1, N]} \big[||\mathbf{x}_{0} -  G(\mathbf{x}_{i}, i)||^2_{2}  \big] \label{eq:diffusion_noise_obj}.
\end{equation}

\subsection{Cross-Entropy Method}
\label{sec:cem}
The CEM finds solutions to complex problems by iteratively refining a probability distribution, often modelled by a Gaussian distribution, to focus on promising solution areas. 
The CEM samples a population of solutions from a given distribution, evaluates them using predefined cost function, and selects the top performing solutions to update the distribution.  
In particular, when using a Gaussian distribution, the mean $\mu$ and variance $\sigma^2$ are updated by computing the mean and variance of the top performing samples.

%% file: sections/approach.tex
\section{Latent Diffusion Trajectory Optimisation}
Given a representative dynamics model (e.g. a simulator or a learned dynamics model), \ourmethod aims to find an action trajectory $\{\mathbf{a}^{0}, \mathbf{a}^{1}, \dots, \mathbf{a}^{T}\}$, over time horizon $T$, that enables a dexterous robot hand to manipulate deformable objects to match a pre-defined goal shape.
\ourmethod consists of three components: a variational autoencoder (VAE) that learns a skill-latent, $\mathbf{z} \in \mathcal{Z}$, encoding short-horizon action trajectories; a latent diffusion model (LDM) that generates sequences of skill-latents that represent entire trajectories for exploration in the state space; and trajectory optimisation using the CEM within the reverse diffusion process for a target task.
\ourmethod relies on a task-agnostic play dataset of action trajectories that cover a wide range of meaningful robot hand motions to learn a skill-latent space and to train the LDM.
The play dataset trajectories, including movements such as closing and opening the hand and moving individual fingers, are collected by mimicking human hand motions. 
The VAE learns the skill-latent space by reconstructing short-horizon action trajectories randomly sampled from the dataset.
Then, the LDM is trained to compose skills to form a long-horizon trajectory that corresponds to a long-horizon action trajectory seen in the dataset.
With the LDM's capability of modelling a complex multi-modal distribution, \ourmethod generates diverse long-horizon trajectories of meaningful motions that allow efficient exploration in the state space.

To find performant trajectories using the LDM for a given task, we propose a novel gradient-free guided sampling method that employs the CEM within the reverse process.
For each reverse step, the LDM generates a small number of noisy trajectories which are a sequence of latent skill representations. 
While the generated sequence of skills is noisy, each short-horizon of actions decoded from the skill representations using the VAE decoder represents meaningful behaviour that facilitates exploration.
After evaluating the sampled trajectories in a simulator, the trajectory with the lowest cost is chosen for further denoising using the LDM, removing noise and resulting in an increasingly minimised task-informed cost. 

\begin{figure*}
    \centering
    \includegraphics[width=1.0\textwidth]{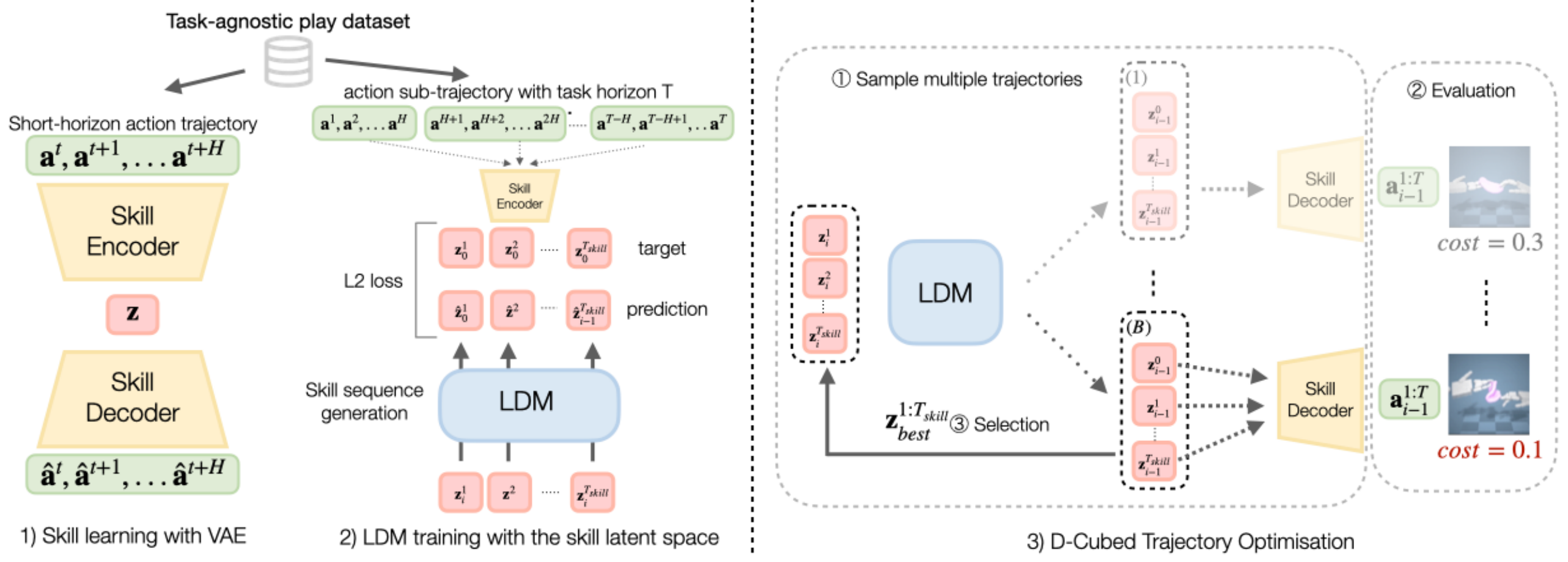}
    \caption{\textbf{Method overview.} (1) A VAE is trained to learn a skill latent representation $\mathbf{z}$ by reconstructing a short-horizon action sequence $\mathbf{a}^{t:t+H}$ randomly sampled from the task-agnostic play dataset. (2) A latent diffusion model (LDM) is trained to compose skills into a skill trajectory, representing a long-horizon action trajectory sampled from the dataset.
    (3) During trajectory optimisation, the LDM generates $B$ skill trajectories $\{\mathbf{z}^{1:T_{skill}}_{i}\}^{|B|}$, where $T_{skill}=\frac{T}{H}$ is the length of skill trajectories. These trajectories are evaluated in a simulator, and the best sequence $\mathbf{z}^{1:T_{skill}}_{best}$, characterised by achieving the minimum cost, is selected for the subsequent reverse process. For further details, see Algorithm~\ref{alg:D-Cubed}.}
    \label{fig:method}
    \vspace{-0.2cm}
\end{figure*}

In this section, we first introduce the data collection process in Section~\ref{sec:data_collection}, describe the latent diffusion model in Section~\ref{sec:skill_diffusion}, and detail the sampling method in Section~\ref{sec:traj_opt}. An overview of the method can be seen in Figure~\ref{fig:method}.

\subsection{Data Collection}
\label{sec:data_collection}

Several works~\cite{li2023dexdeform, Ze2024DP3} require expert demonstrations for each new task, however, this is considered expensive due to the difficulty in manipulating deformable objects through teleoperation systems.
To alleviate this issue, a single task-agnostic play dataset of robot hand trajectories $\mathcal{D}_{play}$ is collected per robot platform, allowing its reuse across multiple tasks.
This play dataset is designed to span the space of meaningful hand motions that will form a skill latent space learnt by a VAE, which includes motions such as closing and opening the hand, moving individual fingers and moving plus flexing the wrist throughout the workspace of the robot.
Since collecting the play dataset does not require manipulating deformable objects unlike expert demonstrations, the play dataset is readily collected by a human operator without any training.
The dataset is collected in 20 minutes by tracking the motion of the human and re-targetting the human hand pose to a robot hand, inspired by prior work~\cite{handa2020dexpilot}.
For further details, see Appendix~\ref{appendix:data_collection}.

\subsection{Latent Diffusion Model as Skill Sampler}
\label{sec:skill_diffusion}
Long-horizon dexterous deformable manipulation tasks induce a large solution space due to the task complexity often caused by numerous contacts with deformable objects and the high DoF of a robot hand.
In such scenarios, sampling low-level actions to find performant solutions is often infeasible because the sampled action trajectory is unlikely to correspond with meaningful robot hand motions that effectively explore the solution space.
Instead, inspired by previous work~\cite{pertsch2021accelerating}, \ourmethod learns a skill latent space that encodes a short-horizon action trajectory from the play dataset, which plays a significant role in efficiently exploring the search space of tasks.
In particular, a VAE, consisting of an encoder $q^{enc}_{\psi}(\mathbf{z}|\mathbf{a}^{t:t+H})$ and a decoder $p^{dec}_{\psi}(\mathbf{a}^{t:t+H}|\mathbf{z})$, is trained to reconstruct short-horizon action trajectories $\mathbf{a}^{t:t+H}$ randomly sampled from the play dataset $\mathcal{D}_{play}$ to learn the skill $\mathbf{z} \in \mathcal{Z}$, by optimising the ELBO:
\begin{equation}
    \mathcal{L}^{\mathrm{ELBO}}=\mathbb{E}_{\mathbf{z} \sim q_\psi(\mathbf{z} \mid \mathbf{a}^{t:t+H})} \log p^{dec}_\psi(\mathbf{a}^{t:t+H} \mid \mathbf{z}) \nonumber -D_{\mathrm{KL}}\left[q_\psi(\mathbf{z} \mid \mathbf{a}^{t:t+H }) \| p(\mathbf{z})\right]
\end{equation}
where $p(\mathbf{z})$ is a Gaussian prior over the latent representation and $H$ is the length of short-horizon action sequence ($H=10$).
See Appendix~\ref{appendix:vae_training} for further details on the architecture and hyperparameters of the VAE.

Since the skill latent representation only encodes short-horizon actions of the robot hand, composing multiple skill latent representations into a skill trajectory that corresponds to a long-horizon action trajectory is necessary for efficient exploration (e.g. to make meaningful contacts with an object).
To achieve this, an LDM is trained to compose a sequence of skill-latent representations $\mathbf{z}^{1:T_{skill}}$, where $T_{skill} = \frac{T}{H}$ is the length of the skill trajectory, which reconstruct robot hand trajectories from the play dataset. 
The LDM is trained to optimise the following objective:

\begin{equation}
    \mathcal{L}_{\text{ldm}} = \mathbb{E}_{\mathbf{z}^{1:T_{skill}}\sim \mathcal{D}_{play}, i\sim[1, N]} \big[||\mathbf{z}_{0}^{1:T_{skill}} -  G(\mathbf{z}_{i}^{1:T_{skill}}, i)||^2_{2}  \big] \label{eq:diffusion_noise_obj}.
\end{equation}

An LDM is chosen in this case as it is a generative model proven to be capable of representing a complex multimodal distribution, like that of the play dataset~\cite{chen2023playfusion}.
In this work, we employ a transformer architecture as the backbone of the noise prediction model $\epsilon_{\theta}(\cdot, i)$ (see Appendix~\ref{appendix:ldm_training} for further details).

\subsection{Trajectory Optimisation using Gradient-Free Guided Sampling}
\label{sec:traj_opt}

Traditional trajectory optimisation often struggles to solve dexterous deformable object manipulation tasks due to the large search space, as mentioned in Section~\ref{sec:skill_diffusion}.
This is further amplified by the limited global task information available from a cost function.
Using the capability of the LDM with skill-latent space, the LDM generates diverse skill trajectories that represent long-horizon action trajectories of meaningful robot hand motions (Section~\ref{sec:skill_diffusion}), leading to effective exploration of the solution space.
However, to solve a target task, guidance is required to direct the diffusion sampling process to converge towards high-performing trajectories.
While classifier guidance is a common technique for guiding the reverse process using gradients, inaccurate~\cite{chao2022denoising} or noisy gradients such as those obtained from differentiable physics simulators~\cite{antonova2022rethinking} are unable to successfully guide the reverse diffusion process (see Section~\ref{sec:main_res} for experimental results).
To avoid issues related to gradients, we propose gradient-free guided sampling that employs the CEM~\cite{rubinstein1999cem} within the reverse diffusion process to optimise a trajectory for a target task.
The reverse process can be viewed as analogous to the CEM optimisation steps, as \ourmethod evaluates generated action trajectories in a simulator and updates the parameters of a Gaussian distribution based on the trajectory with the lowest cost for each diffusion step (see Figure~\ref{fig:method} and Algorithm~\ref{alg:D-Cubed}).

In particular, for each reverse step, a small number of noisy skill trajectories $\mathbf{z}^{1:T_{skill}}$ are sampled from a Gaussian distribution with a mean $\boldsymbol{\mu}_{i}$ predicted by the LDM  (Line~\ref{algline:3},~\ref{algline:10}), where $T_{skill}=\frac{T}{H}$ represents the horizon length of the skill-latent representations.
Crucially, during the early stages of the reverse process, \ourmethod focuses on exploring the search space. 
This is because the variance $\Sigma_{\boldsymbol{\theta}}$ of the Gaussian distribution, determined by the noise scheduling (see Equation~\ref{eq:mean}), is large, thereby generating diverse trajectories.
During later steps of the reverse process, \ourmethod attempts to refine the trajectories for a target task due to the small variance of the distribution.
The generated skill trajectories are decoded using the VAE into low-level action trajectories that are then evaluated in the simulator to obtain their respective scores (Line~\ref{algline:5}).
While the generated skill sequences are noisy in terms of their composition, each short-horizon action sequence decoded from the skill latent representations remains smooth, which effectively promotes meaningful trajectories from the search space for efficient exploration.
Similar to how the CEM updates a Gaussian distribution based on the top-performing samples for each optimisation step (see Section~\ref{sec:cem}), \ourmethod also updates a Gaussian distribution to search for more promising solution areas by predicting the mean $\boldsymbol{\mu}_{i}$ using the LDM given the trajectory with the lowest cost $\mathbf{z}_{best}^{1:T_{skill}}$ as input (see Equation~\ref{eq:mean} and Line~\ref{algline:6}):
\begin{equation}
\label{eq:mean}
    \mu_{\theta}(\mathbf{z}_{i}^{1:T_{skill}}, i) = \frac{\sqrt{\bar{\alpha}_{i-1}} \beta_t}{1-\bar{\alpha}_i} G_{\theta}(\mathbf{z}_{best}^{1:T_{skill}}, i)+\frac{\sqrt{\alpha_t}\left(1-\bar{\alpha}_{i-1}\right)}{1-\bar{\alpha}_i} \mathbf{z}_{best}^{1:T_{skill}}
\end{equation}

Although CEM normally requires several top-performing samples to compute the mean and variance of the Gaussian distribution, \ourmethod needs only a single top-performing sample because the mean at the next diffusion step is determined by the prediction of the LDM.
Furthermore, in \ourmethod, the variance is updated based on a noise schedule (see Equation~\ref{eq:diffusion_variance}) for each reverse step.
Intuitively, by choosing the lowest-cost trajectory for each diffusion step, the LDM removes noise from performant noisy trajectories for the subsequent reverse process. This leads to further refinement of the trajectory and minimisation of the cost function by exploring more promising solution areas.

Lastly, the mean $\boldsymbol{\mu}_{i}$ of the Gaussian distribution predicted by the LDM does not necessarily compose a better distribution for the following reverse steps because the LDM may make a poor prediction, leading to trajectory samples that have a higher cost.
To address this issue, inspired by a variant of CEM~\cite{pinneri2021sample, szita2008online}, the mean of the distribution is updated only when the current predicted mean $\boldsymbol{\mu}_{i}$ has a lower cost than the previous mean with the lowest cost $\boldsymbol{\mu}_\text{best}$ (see Line~\ref{algline:8}, \ref{algline:9}).
This optimisation method, when paired with a LDM is empirically shown in Section~\ref{sec:main_res} to generate performant trajectories. 

\begin{algorithm}[t]
    \caption{Differentiable Physics Guided Trajectory Diffusion Optimisation (\ourmethod)}\label{alg:D-Cubed}
    \begin{algorithmic}[1]
    \State \textbf{Require}: denoising model, $G_{\boldsymbol{\theta}}$; gradient scale, $\delta$; target state of deformable objects, $\mathbf{s}_{\text{target}}$, $T_{skill} = \frac{T}{H}$
    \State \textbf{Initialise:} $C_{\text{best}}=\infty, \boldsymbol{\mu}_{\text{best}} = \text{None}$
    \label{algline:2}
    \vspace{1mm} \label{algline:init_state}
    \State $\{\mathbf{z}^{1}_{N}, \dots, \mathbf{z}^{T_{skill}}_{N}\}^{|B|} \sim \mathcal{N}(\mathbf{0}, \mathbf{I})$ \label{algline:3} \Comment{Sample $B$ initial sequences of latent representations}
    \For{$i =  N, N-1, \ldots, 1$} \label{algline:4}
        \State $\mathbf{z}^{1:T_{skill}}_{best} \leftarrow  \Call{FindBestLatents}{\{\mathbf{z}^{1}_{i}, \dots, \mathbf{z}^{T_{skill}}_{i}\}^{|B|}}$ \Comment{Choose the best sequence of skill latents} \label{algline:5}
        \State $\boldsymbol{\mu}_{i} \leftarrow \boldsymbol{\mu}_{\theta}(\mathbf{z}_{best}^{1:T_{skill}})$ \Comment{Predict a mean of a Gaussian distribution (see Eq.~\ref{eq:latent_mean})} \label{algline:6}
        \State $\text{cost} = \text{evaluate}(q^{dec}_{\theta}(\mathbf{a}^{1:T} | \boldsymbol{\mu}_{i}))$ \Comment{Evaluate the predicted mean}\label{algline:7}
        \If{$\text{cost} < \text{C}_{\text{best}}$} \label{algline:8}
            \State $\boldsymbol{\mu}_{\text{best}} \leftarrow \boldsymbol{\mu}_{i}$, $\text{C}_{best} \leftarrow \text{cost}$ \label{algline:9}
        \EndIf

        \State $\{\mathbf{z}^{1}_{i-1}, \dots, \mathbf{z}^{T_{skill}}_{i-1}\}^{|B|} \sim \mathcal{N}(\boldsymbol{\mu}_{best}, \mathbf{\sigma}_{i-1}^{2}\mathbf{I})$ \Comment{Sample a batch $B$ of a sequence of skill latents}\label{algline:10}
    \EndFor \\
    \Return $p_{\psi}^{dec}(\mathbf{a}^{1:T} | \mathbf{\mathbf{\mu}_{best}})$\label{algline:12} \vspace{1mm}
    
    \Function{FindBestLatents}{$\{\mathbf{z}^{1}, \dots, \mathbf{z}^{T_{skill}}\}^{|B|}$}
        \State $\text{cost}_{\text{best}}, \mathbf{z}_{\text{best}} = \infty, \text{None}$
        \For{$\mathbf{z}^{1:T_{skill}} = \{\mathbf{z}^{1}, \dots, \mathbf{z}^{T_{skill}}\}^{|B|}$}   \Comment{Evaluate each sequence of latent representations in the batch}
            \State $\text{cost}_{j} = \text{evaluate}(p_{\psi}^{dec}(\mathbf{a}^{1:T} | \mathbf{z}^{1:T_{skill}}))$
            \If{$\text{cost} < \text{cost}_{best}$}
                \State $\text{cost}_{\text{best}} \leftarrow \text{cost}$, $\mathbf{z}_{\text{best}} \leftarrow \mathbf{z}^{1:T_{skill}}$
            \EndIf
        \EndFor
        \Return $\mathbf{z}_{\text{best}}$
    \EndFunction
    \end{algorithmic}
    \end{algorithm}

%% file: sections/experiments.tex
\section{Experiments}
Our experimental evaluation is designed to answer the following questions:
(1) How effective is D-Cubed in generating trajectories for dexterous deformable object manipulation?
(2) How does our method compare to competitive baselines including traditional trajectory optimisation approaches and other methods that do not require expert demonstrations?
(3) How does the performance of our method improve during the diffusion process?
(4) How important are the design decisions of \ourmethod in generating high-performance trajectories?

\subsection{Experimental Setup}
\textbf{Simulated Environments}: We evaluate \ourmethod on a suite of six challenging dexterous deformable object manipulation tasks introduced in prior work~\cite{li2023dexdeform}.
The benchmark consists of three single-hand tasks (\emph{Folding}, \emph{Flip}, \emph{Wrap}) and three dual-hand tasks (\emph{Dumpling}, \emph{Bun}, \emph{Rope}).
For dual-arm tasks, \ourmethod samples an individual trajectory for each arm.
The cost function defined by the benchmark is the Sinkhorn Divergence~\cite{sejourne2019sinkhorn} which measures the difference between the manipulated and the target object shape using point clouds.
Our experimental setup closely follows that of prior work~\cite{li2023dexdeform} in that we evaluate \ourmethod and competitive baselines on tasks intended to form $5$ different target shapes for each task.
For more details on the tasks, see Appendix~\ref{appendix:environment} and the prior work~\cite{li2023dexdeform}.

\textbf{Evaluation metric}: Following~\cite{li2023dexdeform}, we report the normalised improvement in the Earth-Mover distance (EMD) approximated by the Sinkhorn Divergence, calculated as $d(t) = \frac{d_{0} - d_{t}}{d_{0}}$ where $d_{0}$ and $d_{t}$ are the initial and current values of the EMD. 
Thus, when the normalised improvement is 1, the shape of the deformable object perfectly matches that of the target object.
A negative score caused by a large discrepancy between the current and the target shape is set to 0.

\subsection{Baselines}
We compare \ourmethod with the following state-of-the-art and competitive baselines that represent competing approaches capable of generating trajectories (for further details, see Appendix~\ref{appendix:baseline}):
\begin{itemize}[leftmargin=5.5mm]
    \item {\bf Grad TrajOpt}: A gradient-based trajectory optimisation~\cite{huang2021plasticinelab} that utilises the first-order gradients available from the benchmark simulator.
    \item {\bf MPPI}: A sampling-based trajectory optimisation method~\cite{williams2017mppi} that samples a batch of short-horizon trajectories from a Gaussian distribution and updates the parameters of the distribution based on the top-performing trajectories.
    \item {\bf Skill-based MPPI}: Skill-based MPPI is similar to the MPPI baseline, but the action space is the skill-latent space.
    In contrast to \ourmethod, which uses an LDM to sample meaningful skill compositions, \emph{Skill-based MPPI} must optimise such meaningful compositions by sampling diverse trajectories. This baseline highlights the importance of the skill composition learnt by the LDM in \ourmethod as a means of exploring the large search space. 
    \item {\bf PPO}: Proximal Policy Optimisation (PPO)~\cite{schulman2017proximal}, a model-free RL method, which takes a point cloud as input. RL is an alternative method to trajectory optimisation and is capable of generating a sequence of actions in a closed-loop.
    \item {\bf LDM w/ Classifier Guidance}: Using the learnt LDM to optimise a trajectory through the reverse process with classifier guidance using gradients from the simulator. This baseline demonstrates that classifier guidance using the noisy gradients available from the simulator struggles to guide the sampling process for trajectory optimisation.
    \item {\bf Diffusion-ES}: A concurrent method~\cite{yang2024diffusiones} that also proposes a gradient-free sampling method based on evolutionary search with a truncated diffusion process. 
\end{itemize}

\begin{figure*}
    \centering
    \includegraphics[width=0.85\textwidth]{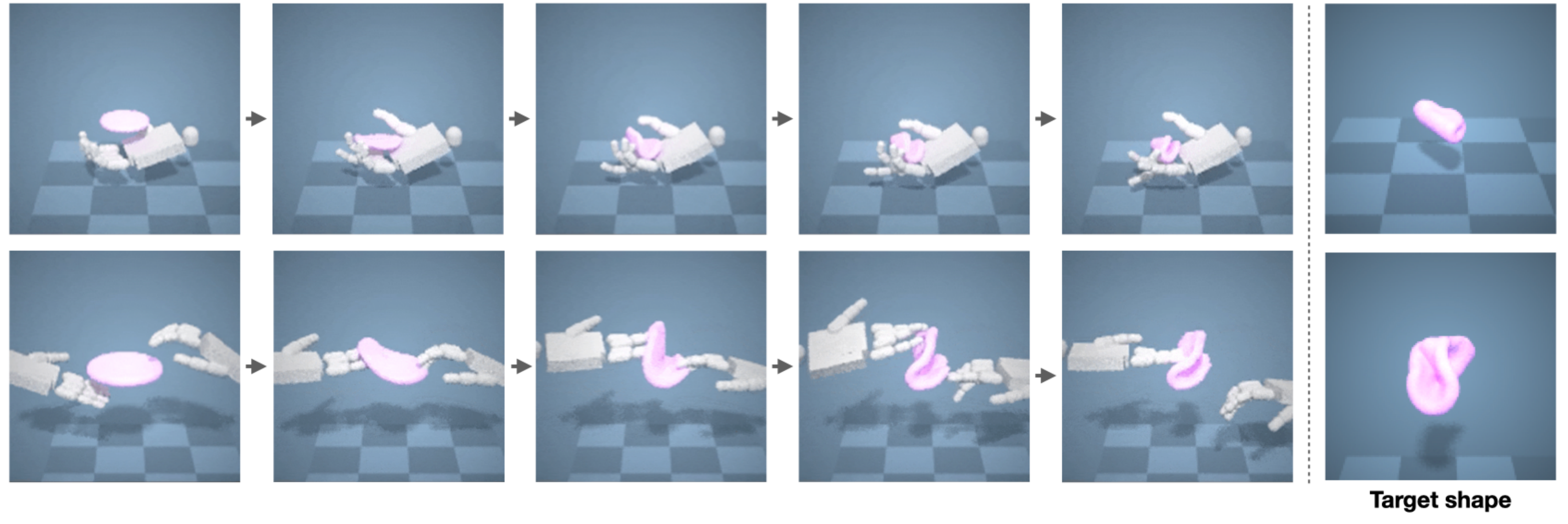}
    \caption{Qualitative results of \ourmethod. (Top) Flip task - the hand, using primarily the wrist and finger DoFs, is able to fold the plasticine into a configuration that is representative of the goal state. (Bottom) Dumpling task - Using two hands to deform the stationary plasticine, \ourmethod is able to manipulate the plasticine close to the target shape.}
    \label{fig:result}
\end{figure*}

\begin{table}[t]
    \centering  
    \scalebox{0.8}{
    \begin{tabular}{c|c|c|c | c | c| c}
        \hline
        \bf{Env} & Folding & Rope &  Bun & Dumpling & Wrap &  Flip \\
        \hline
        Grad TrajOpt & $0.032 \pm 0.061$ & $0.079 \pm 0.026$ & $0.000 \pm 0.000$  & $0.032 \pm 0.061$ & $0.079 \pm 0.026$ & $0.000 \pm 0.000$ \\
        MPPI & $0.002 \pm 0.005$ & $0.000 \pm 0.000$  &  $0.000 \pm 0.000$ & $0.021 \pm 0.042$  &  $0.000 \pm 0.000$  & $0.000 \pm 0.000$ \\
        Skill-based MPPI & $0.020 \pm 0.002$ & $0.000 \pm 0.000$ & $0.048 \pm 0.012$ & $0.052 \pm 0.034$  & $0.000 \pm 0.000$ & $0.000 \pm 0.000$  \\
        PPO & $0.361 \pm 0.173$ & $0.460 \pm 0.257$ & $0.069 \pm 0.117$ & $0.009 \pm 0.014$ & $0.016 \pm 0.016$ & $0.448 \pm 0.080$  \\
        LDM w/ Classifier guidance & $0.050 \pm 0.038$ & $0.001 \pm 0.001$ & $0.019 \pm 0.018$ & $0.009 \pm 0.014$ & $0.016 \pm 0.016$ & $0.448 \pm 0.080$  \\
        Diffusion-ES & $0.403 \pm 0.227$ & $0.192 \pm 0.059$ & $0.273 \pm 0.092$ & $0.179 \pm 0.057$ & $0.305 \pm 0.007$ & $0.678 \pm 0.032$  \\
        D-Cubed & $\mathbf{0.871 \pm 0.021}$ & $\mathbf{0.741 \pm 0.031}$ & $\mathbf{0.704 \pm 0.012}$ & $\mathbf{0.699 \pm 0.037}$ & $\mathbf{0.512 \pm 0.032}$ & $\mathbf{0.909 \pm 0.025}$
    \end{tabular}}
    \vspace{0.3cm}
    \caption{The averaged normalised improved EMD and standard deviation is reported for each method. The scores are averaged over $3$ seeds. The scores for \emph{Grad TrajOpt} and \emph{PPO} are taken from previous work~\cite{li2023dexdeform}.}
    \label{tab:main_result}
\end{table}

\subsection{Trajectory Optimisation Results}
\label{sec:main_res}
Table~\ref{tab:main_result} shows the normalised improvement in EMD for each task.
As shown in Table~\ref{tab:main_result}, \ourmethod outperforms the baseline methods by a significant margin on all tasks.
The results indicate that \ourmethod effectively combines the learnt skills to search the solution space using the LDM, and exploits the diverse sampled trajectories using the gradient-free guided sampling to minimise a task-informed cost.
Traditional trajectory optimisation baselines, including \emph{Grad TrajOpt} and \emph{MPPI}, struggle to find performant trajectories because such methods fail to explore the action space effectively and rarely receive useful task information from the cost function to update the sequence of actions.
Moreover, \emph{Skill-based MPPI} that utilises a skill-latent space as an action space in \emph{MPPI} also does not perform well.
This emphasises the effectiveness of using an LDM, which is trained to generate meaningful sequences of skills, restricting the search space to only meaningful trajectories for exploration.
\emph{PPO} performs well compared to \emph{Grad TrajOpt} on several tasks because \emph{PPO} can explore the action space more effectively by sampling actions from a stochastic policy.
However, \emph{PPO} does not achieve competitive performance compared to \emph{D-Cubed} because RL often struggles to explore large state spaces.

\emph{LDM w/ Classifier guidance}, which utilises gradients from the simulator, performs poorly compared to \ourmethod on all tasks.
The noisy gradients obtained from the simulator~\cite{antonova2022rethinking} and the limited task information available in the gradients, especially when there is no contact between the robot hand and objects, hinder the guidance within the reverse process.
This issue is particularly prominent in tasks that require extensive search and contact with objects.

\emph{Diffusion-ES} also struggles to achieve competitive performance in the suite of challenging benchmark tasks.
\emph{Diffusion-ES} initially generates clean trajectories and repeats the forward and reverse processes of short diffusion steps to perturb the trajectories with lower costs.
However, due to the small perturbations added to the trajectories, it does not effectively explore the search space, particularly when the initially sampled trajectories do not make contact with the objects.
On the other hand, \ourmethod effectively explores the solution space at the beginning of the reverse diffusion process by generating noisy skill trajectories.
Crucially, while Diffusion-ES generates low-level action trajectories, \ourmethod generates skill trajectories, allowing for more efficient exploration in the state space.

\subsection{Ablation Studies}
\label{sec:ablation}

\begin{figure*}[t!]
\begin{minipage}{\textwidth}
  \centering
  \begin{subfigure}[t]{0.48\textwidth}
    \centering
    \includegraphics[width=\textwidth]{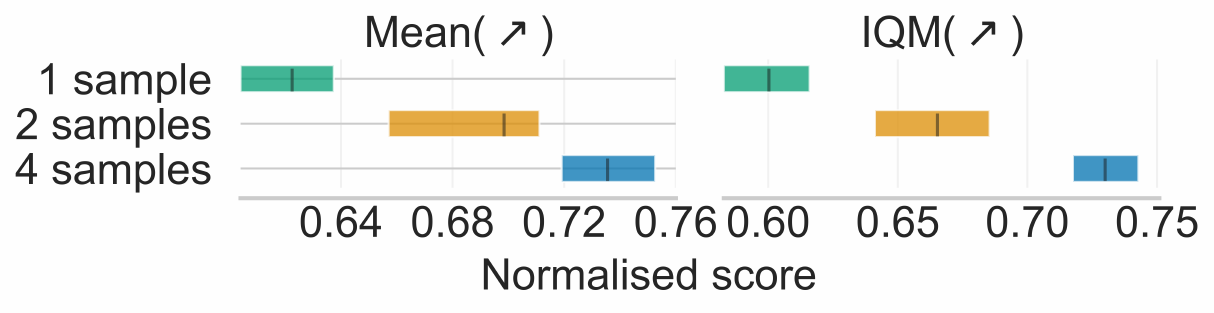} 
    \caption{Number of sampled trajectories}
    \label{fig:ablation_num_samples}
  \end{subfigure}
  \begin{subfigure}[t]{0.48\textwidth}
      \centering
      \includegraphics[width=\textwidth]{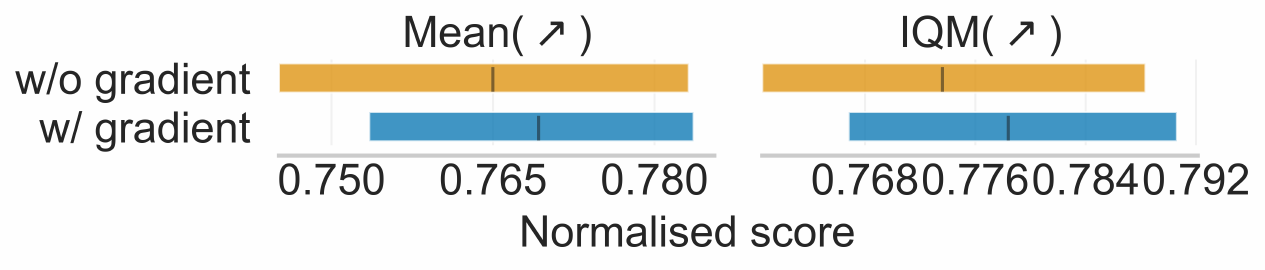}
      \caption{Additional gradient guidance}
      \label{fig:ablation_grad_samples}
  \end{subfigure}
\end{minipage}
\caption{
We report Mean and Interquartile Mean (IQM) of improvement in EDM averaged across all six tasks.
(a) Ablation of the number of trajectories sampled in our proposed gradient-free guided sampling (line~\ref{algline:10} in Algorithm~\ref{alg:D-Cubed}). (b) Comparison of performance with and without additional gradient guidance in our method.
}
\vspace{-1em}
\end{figure*}

\begin{wrapfigure}{r}{0.4\textwidth}
    \vspace{-1cm}
    \centering
    \includegraphics[width=0.4\textwidth]{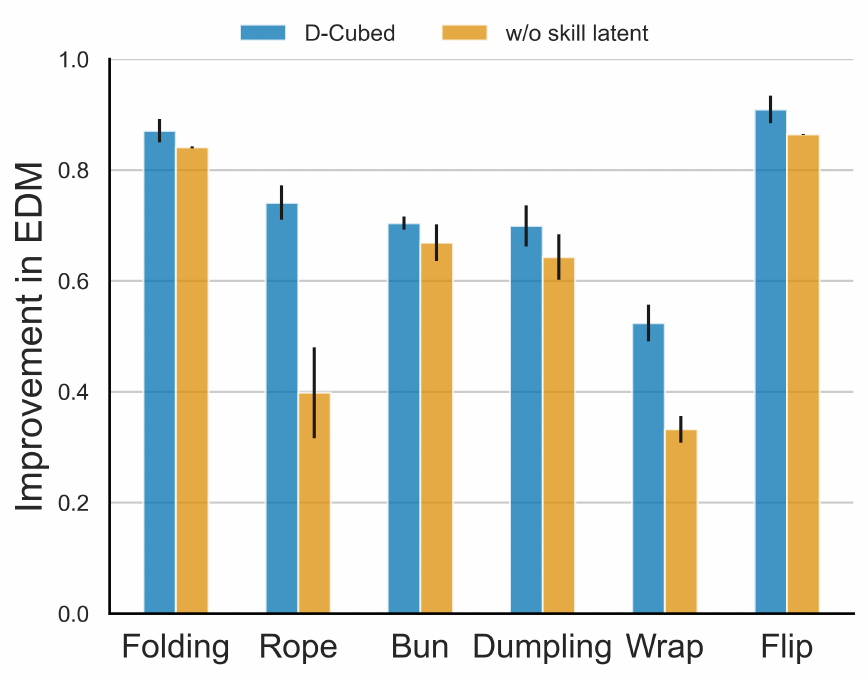}
    \caption{Comparison of \ourmethod w/ and w/o skill latent representations.}
    \label{fig:ablation_skill}
    \vspace{-0.4cm}
\end{wrapfigure}

We investigate the importance of different design decisions of \ourmethod.
Fig.~\ref{fig:ablation_num_samples} shows the mean and  Interquartile Mean (IQM) with $95\%$ confidence intervals (CIs) of normalised improvement in EMD with different numbers of trajectories sampled during the reverse diffusion process (Line~\ref{algline:10}  in Algorithm~\ref{alg:D-Cubed}), following the evaluation method of~\cite{agarwal2021deep}.
As shown in Fig.~\ref{fig:ablation_num_samples}, sampling more trajectories during the reverse process significantly improves performance because \ourmethod can more effectively search the solution space.

The performance disparity when \ourmethod also uses classifier guidance with the proposed gradient-free sampling is also reported in Fig.~\ref{fig:ablation_grad_samples}.
As classifier guidance does not demonstrate a statistically significant improvement in the score compared to \ourmethod without gradient guidance, the increased time required by the simulator to calculate gradients used for gradient guidance is not warranted. Additionally, this result adds further evidence to the premise that gradients from differentiable simulators are often sparse and uninformative~\cite{antonova2022rethinking}.

Finally, whether the pre-trained skill-latent space improves task performance is assessed.
Fig.~\ref{fig:ablation_skill} shows the performance of \ourmethod with and without the use of a skill latent across the six tasks. It is evident in the simpler tasks that do not require extensive searches of the solution space, such as \emph{Flip} and \emph{Folding}, that the use of a skill-latent space does not dramatically improve results.
However, the performance of \ourmethod on \emph{Rope} and \emph{Wrap}, considerably harder tasks, is substantially better when using a skill-latent space.
We reason that this is because \ourmethod when using a skill latent space can effectively search the solution space and tackle hard-exploration problems.

\subsection{Qualitative Results in Real-World Environments}
\begin{figure*}[t]
    \centering
    \includegraphics[width=0.9\textwidth]{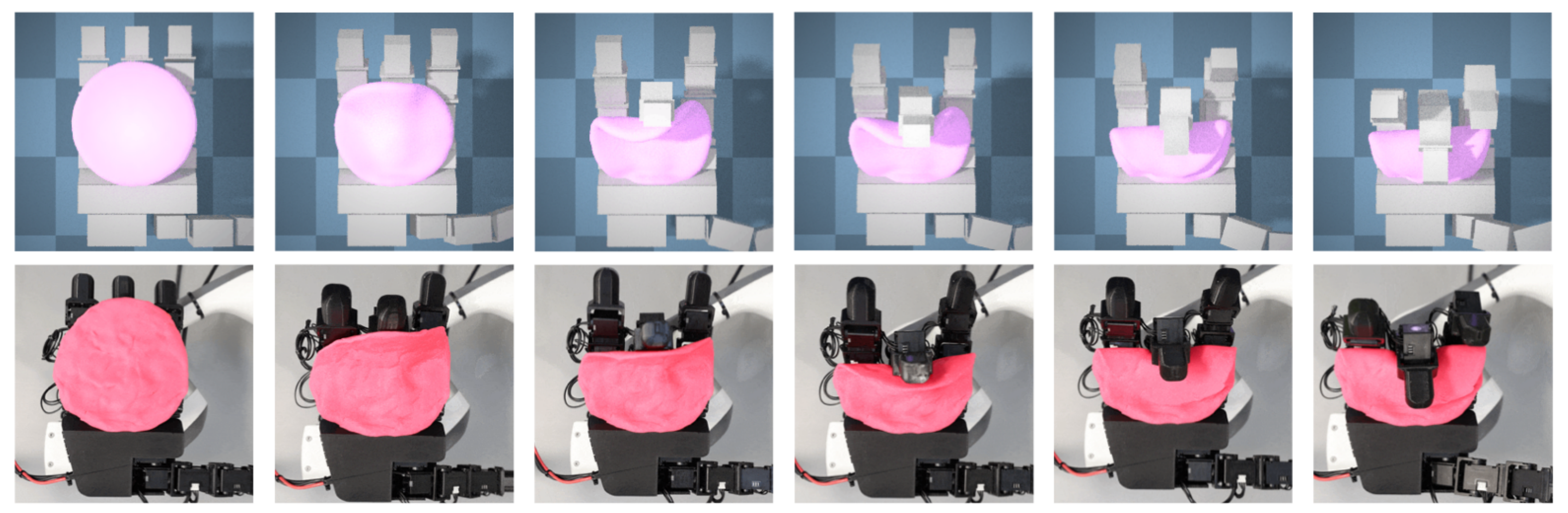}
    \caption{Qualitative results of \ourmethod using the LEAP hand in a real-world experiment. The LEAP hand effectively manipulates the deformable object, exhibiting similar deformation as observed in the simulation.}
    \label{fig:real_result}
\end{figure*}
We qualitatively investigate whether an optimised trajectory from simulation can be transferred to real-world environments.
Due to hardware limitations, we use a LEAP hand, a low-cost dexterous hand~\cite{shaw2023leaphand}, instead of the Shadow hand~\cite{shadowhand}.
Thus, the simulator is modified by replacing the Shadow hand with the LEAP hand.
We choose to evaluate the trajectory transfer on the \emph{Flip} task, as this benchmark task makes the least number of simplifying assumptions compared to the real world.
The other tasks permit object interpenetration with the table and unrealistic floating behaviour of the deformable object, rendering the evaluation impractical. %
In this experiment, we assume that the start state of the deformable object is known.
Thus, we transfer the sequence of actions optimised in the simulated environment to the real-world environment and control the hand in an open-loop manner.
As shown in Fig.~\ref{fig:real_result}, the hand successfully \emph{flips} the deformable object so as the object is folded within the hand in the real-world environment.

\subsection{Limitations}
Experimental studies demonstrate that \ourmethod can generate a performant action sequence for dexterous deformable object manipulation tasks. 
However, \ourmethod is unable to find trajectories that would transfer to the real world for all tasks, because some of the benchmark tasks allow for non-realistic behaviour such as penetration of the table and floating of deformable objects. 
Limitations of \ourmethod include the time to generate a desired trajectory, which is heavily dependent on the time required to evaluate trajectories in a simulation. 
However, using a faster, parallel simulator~\cite{makoviychuk2021isaac} instead of the slow, non-parallel simulator from prior work \cite{li2023dexdeform} would greatly improve the speed of optimising a trajectory. 
Finally, \ourmethod is open-loop and therefore is unable to accommodate for discrepancies observed when executing the trajectory. 
In future, we will look to close the loop, potentially by distilling the trajectories within a policy.

%% file: sections/conclusion.tex
\section{Conclusion} 
In this work, we present \emph{D-Cubed}, a new trajectory optimisation method to solve long-horizon dexterous deformable object manipulation tasks using a latent diffusion model trained from a task-agnostic play dataset. 
\ourmethod leverages a novel gradient-free guided sampling method that adapts the CEM within the reverse diffusion process.
The proposed sampling method effectively explores the large search space by sampling meaningful robot hand trajectories consisting of a sequence of skill-latent representations and exploits the trajectories to iteratively improve the trajectory performance.
The experimental results show that \ourmethod outperforms the traditional and competitive trajectory optimisation baselines by a significant margin, showing great promise for other challenging trajectory optimisation tasks.

%% file: sections/appendix.tex
\appendix
\section{Additional Analysis}
\label{appendix:additional_analysis}
\subsection{Improvement of Performance over Diffusion Process}
\label{appendix:improvement}
\begin{figure}[h]
    \centering
    \includegraphics[width=0.4\textwidth]{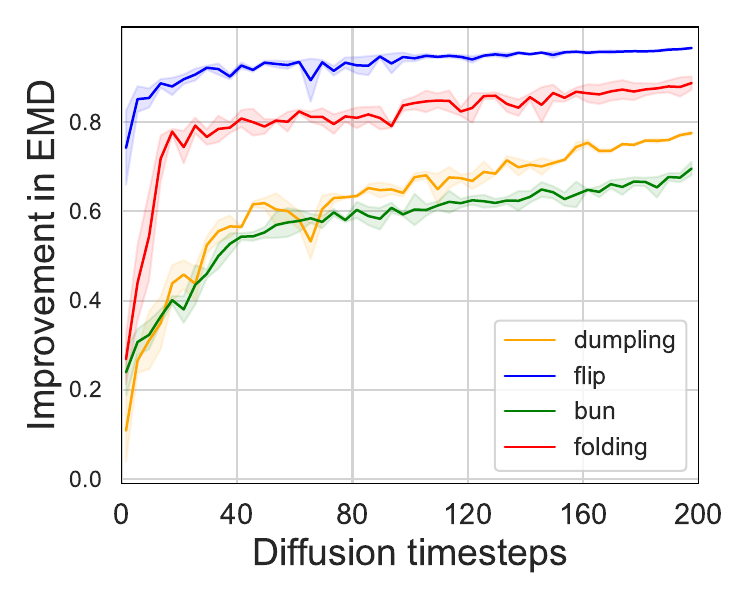}
    \caption{Improvement of EMD during trajectory optimisation in D-Cubed. Our proposed sampling method successfully improves the performance (\ie minimises a cost function) over the diffusion timesteps.}
    \label{fig:improvement}
\end{figure}

We report how our proposed sampling method increases the performance over the diffusion timesteps in Fig.~\ref{fig:improvement}.
As shown in the figure, the normalised improvement in EMD drastically increases at the beginning of the diffusion timesteps, indicating that dexterous robot hands successfully make contact with deformable objects.
Then, our method keeps refining the performance over the diffusion timesteps to minimise a cost function.

\section{Data collection Details}
\label{appendix:data_collection}
A task-agnostic play dataset of representative robot hand motions, including finger closing and opening and wrist movement, is collected.
We use a RealSense D435 camera to track human hand motion and re-target the human hand pose to a robot hand in the SAPIEN simulator~\cite{Xiang_2020_SAPIEN}, inspired by prior work~\cite{handa2020dexpilot}.

We collect the play data for a duration of only 20 minutes, which corresponds to around $50K$ data points.

\section{Training Details}

\subsection{VAE}
\label{appendix:vae_training}
The VAE encoder and decoder consist of a $4$ layer LSTM~\cite{hochreiter1997long} with the dimension of $256$. 
In this work, we use a subsequence of actions with $H=10$ to learn the skill-latent space. 
The VAE is trained using the Adam optimiser~\cite{kingma2014adam} with a learning rate of $1\mathrm{e}{-4}$.

\subsection{Latent Diffusion Models}
\label{appendix:ldm_training}
The latent diffusion model (LDM) used in \ourmethod uses a transformer architecture as the backbone of the noise prediction model.
We use the transformer architecture used in NanoGPT~\footnote{https://github.com/karpathy/nanoGPT}. 
We report further hyperparameter details of the transformer denoiser network and diffusion in Table~\ref{table:transformer_params} and Table~\ref{table:diffusion_params}.

\begin{table}[h]
\centering
\caption{Transformer Denoiser Network Hyperparameter}
\vspace{0.5em}
\begin{tabular}{c|c} 
 \toprule
 Parameter & Value  \\ 
 \midrule
 \midrule
 Optimiser & Adam  \\ 
 Learning rate & 1e-4\\
 Minibatch size & 256 \\
Embedding dimension  & 312 \\ 
Batch size & 256 \\
Number of layers & 6 \\
Self-attention heads & 4 \\ 
 \bottomrule
\end{tabular}
\label{table:transformer_params}
\end{table}

\begin{table}[h]
\centering
\caption{Diffusion Hyperparameters}
\vspace{0.5em}
\begin{tabular}{c|c} 
 \toprule
 Parameter & Value  \\ 
 \midrule
 \midrule
 Number o
 f diffusion timesteps & $200$ \\ 
 Noise schedule & cosine \\
 Noise schedule parameters $s$ & 0.008 \\
 \bottomrule
\end{tabular}
\label{table:diffusion_params}
\end{table}

\section{Baseline Method Details}
\label{appendix:baseline}
As we report the scores for gradient-based trajectory optimisation (TrajOpt) and PPO from prior work~\cite{li2023dexdeform}, we refer the reader to the prior work for further details.

\subsection{MPPI}
\emph{MPPI} baseline samples $30$ trajectories with a horizon of $15$ steps. These parameters are chosen as they result in the same optimisation time as \ourmethod.

\subsection{Skill-based MPPI}
\emph{Skill-based MPPI} baseline samples skill-latent representations for effective exploration of the state space. 
We sample $30$ trajectories with a horizon of $15$ steps that corresponds to a sequence of $150$ actions after decoding the skills.

\subsection{LDM w/ Classifier Guidance}
\emph{LDM w/ Classifier Guidance} baseline leverages classifier guidance~\cite{dhariwal2021diffusion} to generate a desired trajectory.
In particular, first-order gradients from the differentiable physics simulator are used to guide the reverse process of the latent diffusion model.
In our experiments, we denoise a noisy trajectory without classifier guidance for the first half of the diffusion steps so that a relatively clean trajectory can be obtained.
For the rest of the diffusion steps, the following classifier guidance is applied:
\begin{equation}
    \nabla_{\mathbf{x}_{i}} \log p_{\alpha_{i}}(\mathbf{x}_{i} | y) =  \nabla_{\mathbf{x}_{i}} \log p_{\alpha_{i}}(\mathbf{x}_{i}) + \gamma \nabla_{\mathbf{x}_{i}} \log p(y|\mathbf{x}_{i}).
\end{equation}
where $y$ is the cost of the trajectory,  $\nabla_{\mathbf{x}_{i}} \log p(y|\mathbf{x}_{i})$ corresponds to the first-order gradients obtained from differentiable physics simulators, and $\gamma$ is the scale of the classifier guidance.
In our experiment, we use $\gamma=1\mathrm{e}{-4}$.

\subsection{Diffusion-ES}
\emph{Diffusion-ES}, concurrent research~\cite{yang2024diffusiones}, also optimises a trajectory using gradient-free guided sampling with a truncated diffusion process. 
While the prior work chooses the last trajectory of the optimisation process as output, we observe that it is often worse than the trajectories found in the middle of optimisation iterations. 
Thus, we report the score of the best trajectory found during the trajectory optimisation process.

\section{Task Details}
\label{appendix:environment}
\subsection{Cost Function}
The cost function used for trajectory optimisation is defined by Sinkhorn Divegence.
Following the prior work~\cite{li2023dexdeform}, the \emph{geomloss} library is used to define the cost function:
\begin{lstlisting}
from geomloss import SamplesLoss
OT_LOSS = SamplesLoss(loss="sinkhorn", p=1, blur=0.0001)
\end{lstlisting}

\subsection{Tasks}
For single-hand task, such as \emph{Folding} and \emph{Wrap}, the action dimension is 26 (20 for actuators including finger joints and wrist, and 6 for the base). 
For in-hand manipulation tasks (\emph{Flip}), a single hand with a fixed base is assumed, resulting in an action dimension of 20.
In dual-hand environments, the action dimension is 52, allowing for a movable base for both hands.
Since a VAE is trained to encode a single-arm action trajectory in the play dataset, an LDM generates a single-arm skill trajectory.
Thus, to handle dual-hand tasks using \ourmethod the LDM generates a trajectory for each arm.
In the following, we describe the details of each task.

\textbf{Folding}: The initial position of the robot hand is above the dough, and the hand must fold the dough in four different directions: front, back, left, and right.

\textbf{Wrap}: The robot hand first picks up the plasticine ball and places it onto the dough shaped like a rope. Then, it pinches the side of the rope to wrap the ball inside it.

\textbf{Flip}: The robotic hand tosses the dough wrapper in the air to reshape and reposition it.

\textbf{Bun}: The two robotic hands deftly pinch and push the dough to form a bun-shaped object.

\textbf{Rope}: The right-hand grasps the rope on the right, lifts it, and places it above the left rope. Then, the left-hand bends the left rope.

\textbf{Dumpling}: To wrap a dumpling, the right hand first grasps the right side of the wrapper. While holding the dumpling with the right hand, the left hand lifts the left side of the wrapper. Finally, the two hands bring the two sides of the wrapper together and form it into a dumpling shape.